\documentclass[10pt,twocolumn,letterpaper]{article}

\usepackage[pagenumbers]{cvpr} 

\usepackage{graphicx}
\usepackage{amsmath}
\usepackage{amssymb}
\usepackage{booktabs}
\usepackage{makecell}
\usepackage{multirow}

\usepackage[pagebackref,breaklinks,colorlinks]{hyperref}

\usepackage[capitalize]{cleveref}
\crefname{section}{Sec.}{Secs.}
\Crefname{section}{Section}{Sections}
\Crefname{table}{Table}{Tables}
\crefname{table}{Tab.}{Tabs.}

\def\cvprPaperID{1483} 
\def\confName{CVPR}
\def\confYear{2022}

\begin{document}

\title{Decoupling and Recoupling Spatiotemporal Representation for RGB-D-based Motion Recognition}

\author{Benjia Zhou$^{1,2}$\thanks{Work done during an internship at Alibaba Group.}, 
Pichao Wang$^{2}$\thanks{Corresponding author, project lead.}, \\
Jun Wan$^{1,3}$\thanks{Corresponding author.}, Yanyan Liang$^{1}$, Fan Wang$^{2}$, Du Zhang$^{1}$, Zhen Lei$^{3}$, Hao Li$^{2}$, Rong Jin$^{2}$\\
$^{1}$Macau University of Science and Technology \quad $^{2}$Alibaba Group \\
 $^{3}$National Laboratory of Pattern Recognition, Institute of Automation, Chinese Academy of Sciences
}
\maketitle
\begin{abstract}
    Decoupling spatiotemporal representation refers to decomposing the spatial and temporal features into dimension-independent factors. Although previous RGB-D-based motion recognition methods have achieved promising performance through the tightly coupled multi-modal spatiotemporal representation, they still suffer from (i)  optimization difficulty under small data setting due to the tightly spatiotemporal-entangled modeling;
    (ii) information redundancy as it usually contains lots of marginal information that is weakly relevant to classification; and (iii) low interaction between multi-modal spatiotemporal information caused by insufficient late fusion.
    To alleviate these drawbacks, we propose to decouple and recouple spatiotemporal representation for RGB-D-based motion recognition. Specifically, we disentangle the task of learning spatiotemporal representation into 3 sub-tasks: (1) Learning high-quality and dimension independent features through a decoupled spatial and temporal modeling network. (2) Recoupling the decoupled representation to establish stronger space-time dependency. (3) Introducing a Cross-modal Adaptive Posterior Fusion (CAPF) mechanism to capture cross-modal spatiotemporal information from RGB-D data. Seamless combination of these novel designs forms a robust spatialtemporal representation and achieves better performance than state-of-the-art methods on four public motion datasets. Our code is available at https://github.com/damo-cv/MotionRGBD.
\end{abstract}

\begin{figure}[t]
  \centering
   \includegraphics[width=1.0\linewidth]{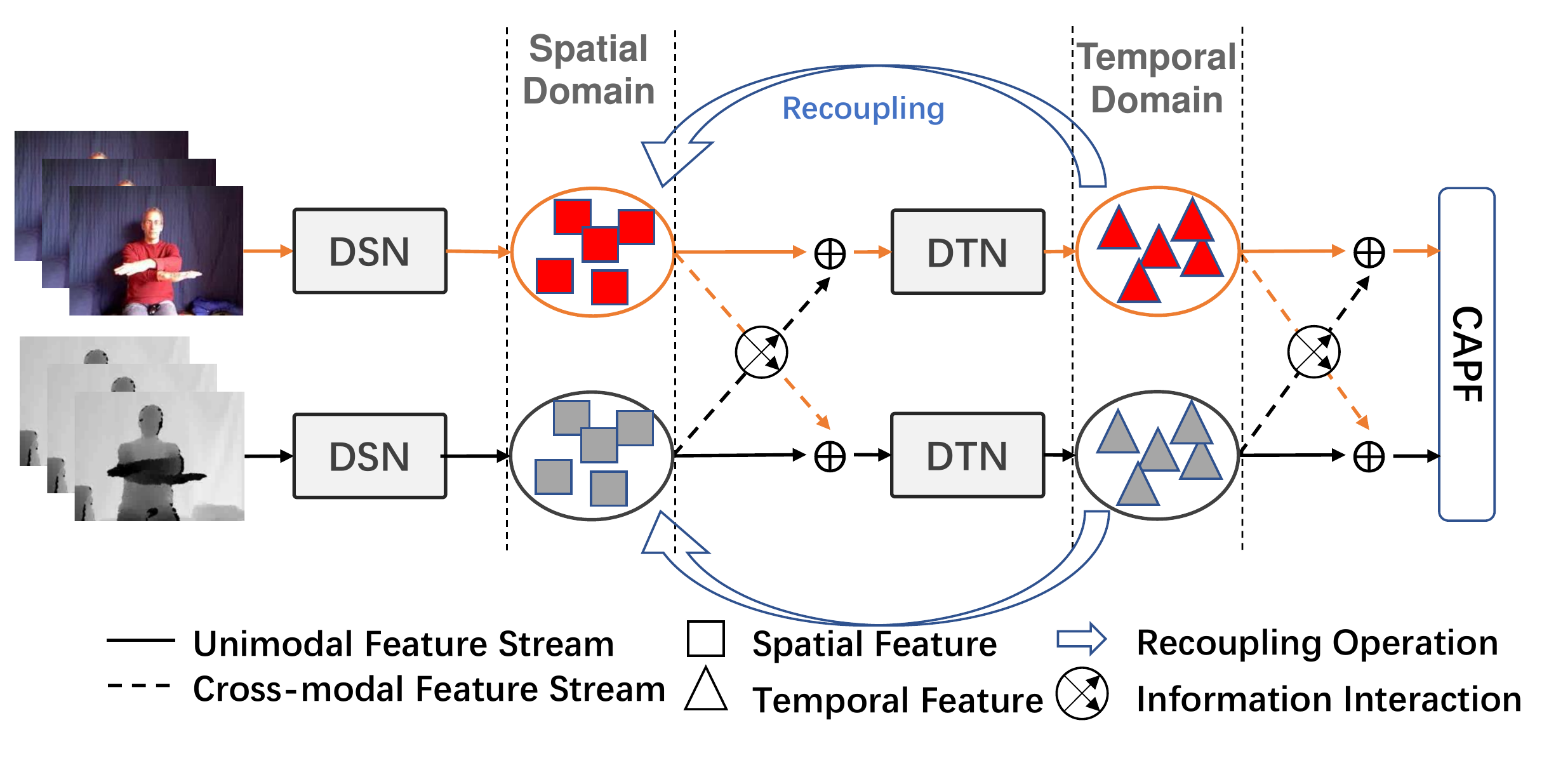}
   \caption{Illustration of the proposed multi-modal spatiotemporal representation learning framework. The RGB-D-based motion recognition can be described as spatiotemporal information decoupling modeling, compact representation recoupling learning, and cross-modal representation interactive learning. Wherein DSN and DTN represent decoupled spatial and temporal representation learning networks, respectively;
}
   \label{fig:CAPFM}
\end{figure}


\section{Introduction}
\label{sec:intro}
Recently, the CNN and RNN based RGB-D motion recognition methods greatly improve the performance of recognition on both gesture \cite{nguyen2019neural, Zhou_Li_Wan_2021, Abavisani_2019_CVPR, yu2021searching, Wang_2017_ICCV} and action \cite{wang2018cooperative, Hu_2018_ECCV, 2018Depth, de2020infrared} through fully exploring the color and depth cues.
Meanwhile, inspired by the transformer scaling success in vision tasks, Transformer-based methods \cite{li2021trear, Eusanio2020gesture} also achieve surprising results on RGB-D-based motion recognition by introducing the cross-attention module for multi-modality fusion. 
\begin{figure*}[t]
  \centering
   \includegraphics[width=1.0\linewidth]{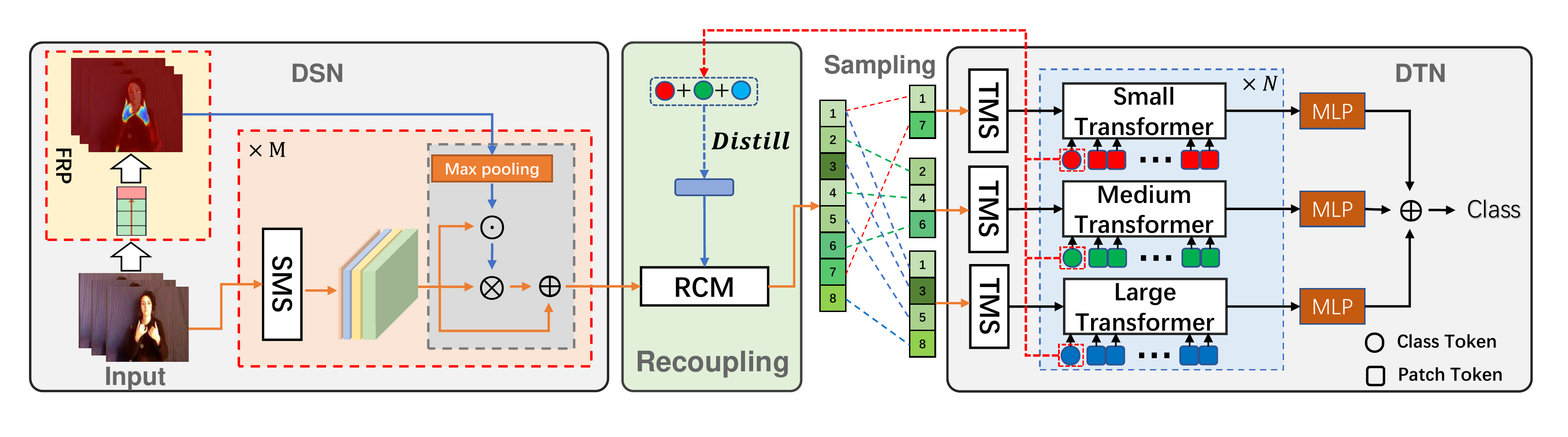}
   \caption{Illustration of proposed decoupling and recoupling saptiotemporal representation learning network. The whole network mainly consists of a decoupled spatial and temporal representation learning networks namely DSN and DTN, as well as a spatiotemporal recoupling module (RCM). The FRP indicates a fast regional positioning module; SMS and TMS indicate the space- and time-centric multi-scale Inception Module respectively. $\bigodot$, $\bigotimes$ and $\bigoplus$ indicate element-wise product, 1D convolution and element-wise add operation respectively.
}
   \label{fig:pipline}
\end{figure*}

Although these works make great progress, we find they are still problematic in the following three aspects. (i) \textbf{Optimization difficulty} exists in the case of limited RGB-D data due to the tightly spatiotmporal entangled modelling.
(ii) \textbf{Redundant information} is hard to deal with in the entangled space-time space. 
To address the above two issues, some decoupled networks (\ie, 2D CNN+LSTM/Transformer\cite{wan2015explore, kalfaoglu2020late}) are proposed to learn the spatiotemporal independent representation. However, we argue that these methods are not conducive to compact representation as they somewhat weaken or even destroy the original spatiotemporal coupling structure. Considering that a certain number of human action classes have strong correlations between time and space, the recoupling process after spatiotemporal decoupling is still necessary. 
(iii) \textbf{Insufficient interaction} occurs between multi-modal spatiotemporal information. Several works \cite{Zhou_Li_Wan_2021, zhu2019redundancy} adopt independent branches for unimodal spatiotemporal representations learning followed by late fusion, resulting in insufficient cross-modal information communication. Thus, it is still a challenge to explore high quality multi-modal spatiotemporal features. 

Given the aforementioned concerns, as illustrated in Figure~\ref{fig:CAPFM}, we introduce a new method of multi-modal spatiotemporal representation learning for RGB-D-based motion recognition. It mainly consists of a decoupled spatial representation learning network (DSN), a decoupled temporal representation learning network (DTN) and a cross-modal adaptive posterior fusion module (CAPF). For each unimodal branch, as shown in Figure~\ref{fig:pipline}, we propose a decoupling and recoupling spatiotemporal feature learning method, wherein a spatiotemporal recoupling module (RCM) is designed as a bonding of DSN and DTN.  RCM acts as feature selection for DSN and knowledge integration for DTN. The entire framework can be decomposed into 3 steps:
(1) \textbf{Spatiotemporal decoupling learning}. In the DSN, 
the video clips are first fed into stack of inception-based spatial multi-scale features learning (SMS) modules to extract hierarchical spatial features. Meanwhile, they are also input into a bypath network, called fast regional positioning module (FRP), to generate visual guidance map, which guides the network to focus on local important areas in the video frame. 
Then the integrated spatial features from SMS and FRP are fed into RCM for feature selection. After that, we sample several sub-sequences at different frame rates from the enhanced spatial features as input to the DTN. The DTN is configured as a multi-branch structure with an inception-based temporal multi-scale layer (TMS) and multiple Transformer blocks for hierarchical local fine-grained and global coarse-grained temporal feature learning. 
(2) \textbf{Spatiotemporal recoupling learning}. To rebuild the space-time interdependence, a self-distillation-based recoupling strategy is developed. As shown by the red dashed line in Figure~\ref{fig:pipline}, the recoupling method is designed as an inner loop optimization mechanism to distill the inter-frame correlations from time domain into the space domain, to
enhance the quality of the spatial features via RCM. 
(3) \textbf{Cross-modal interactive learning}. For multi-modal representation learning from RGB-D data, as shown in Figure~\ref{fig:CAPFM},
we propose an interactive cross-modal spatiotemporal representation learning method. Specifically, the cross-modal spatial features derived from unimodal branches firstly interact at the spatial level and are mapped to a joint spatial representation. Then it is separately integrated with the two unimodal spatial feature streams through the residual structure. After that, the two spatial feature streams are input into their respective temporal modeling networks to capture temporal features. Similar to the spatial feature interaction, a joint temporal representation can also be obtained through interaction at the temporal level. 
Combined with the joint temporal representation, the two temporal feature streams are fed into the CAPF, which is based on a multi-loss joint optimization mechanism, to conduct deep multi-modal representation fusion.

Through the above design, our method not only effectively achieves the spatiotemporal information decoupling and recoupling learning within each modality, but also realizes the deep communication and fusion of multi-modal spatiotemporal information. The proposed method achieves state-of-the-art performance on four public RGB+D gesture/action datasets, namely NvGesture \cite{molchanov2016online}, Chalearn IsoGD~\cite{wan2016chalearn}, THU-READ ~\cite{tang2018multi}, and NTU-RGBD~\cite{shahroudy2016ntu}.

\vspace{-0.4cm}
\section{Related Work}
\label{sec:related}
\subsection{Motion Recognition based on RGB-D Data}
\label{sec:rgbd}
Human motion recognition is one of the most important topics in the field of human-centered research.
With the availability of low-cost RGB-D sensors, RGB-D-based motion recognition has attracted extensive attention. To effectively encode the robust multi-modal spatiotemporal information for motion recognition, Zhu \etal~\cite{zhu2016large} presents a pyramidal-like 3D convolutional network structure for spatiotemporal representation extraction and fusion. 
Kong \etal~\cite{kong2015bilinear} propose to compress and project the RGB-D data into a shared space to learn cross-modal features for effective action recognition. 
Yu \etal~\cite{yu2021searching} employ NAS to search for modal-related network structures and optimal multi-modal information transmission path for RGB-D data. 
Different from the modal-separated multi-branch networks, scene flow is adopted in \cite{wang2017scene} for compact RGB-D representation learning. Wang \etal~\cite{wang2018cooperative} propose to use a single network c-ConvNet for multi-modal spatiotemporal representation learning and aggregation. 
Unlike previous methods that interact with the cross-modal information on coupled spatiotemporal information, we focus on the interaction of multi-modal features in two independent dimensions of space and time.


\subsection{Decoupled Spatiotemporal Feature Learning}
Considering the importance of decoupled spatiotemporal feature learning in sequence,
Shi \etal~\cite{shi2020decoupled} present a decoupled spatiotemporal attention network (DSTA-Net) for skeleton-based action recognition. 
Liu \etal \cite{liu2021decoupled} present a decoupled spatiotemporal Transformer (DSTT) architecture to improve video inpainting tasks. They disentangle attention learning into 2 sub-tasks: (1) attending the temporal object movements on different frames at same spatial locations; (2) attending similar background textures on same frame of all spatial positions.
He \etal~\cite{he2019stnet} propose an effective spatiotemporal network StNet, which employs separated channel-wise and temporal-wise convolution operations for decoupled local and global representation learning.
Zhang \etal~\cite{zhang2020hierarchically} present a hierarchically decoupled spatiotemporal contrastive learning method for self-supervised video representation learning. They capture the spatial and temporal features by decoupling the learning objective into two contrastive sub-tasks, and perform it hierarchically to encourage multi-scale understanding.
In contrast, in this work, we target to learn recoupled features due to strong correlations between time and space for some human actions. Thus, a distillation-based recoupling process is introduced on the decoupled spatiotemporal features.

\section{Proposed Method}
\label{sec:method}
In this paper, we assume the spatiotemporal representation can be decomposed into two sub-domains: the spatial domain that correlates with the visual information, and the temporal domain that describes the time-related concept. Based on this, for unimodal spatiotemporal representation learning, we first decouple the spatiotemporal modeling process to learn domain-independent representations (Sec.~\ref{sec:decoupling}). Then a recoupling method is introduced based on the inner loop optimization mechanism during the training stage (Sec.~\ref{sec:recoupling}), to strengthen the spatiotemporal connection. For multi-modal features interactive learning, we first separately integrate the cross-modal spatiotemporal information into the spatial and temporal domains, and then employ an adaptive posterior fusion mechanism to further fuse the multi-modal features (Sec.\ref{sec:fusion}).

\subsection{Decoupling Spatiotemporal Representation}
\label{sec:decoupling}
\subsubsection{Decoupled Spatial Representation Learning (DSN)}
As shown in Figure~\ref{fig:pipline}, the DSN is composed of fast regional positioning module (FRP) and stack of inception-based spatial multi-scale features learning (SMS) modules. 
Let $[I_1, I_2, \dots, I_T]$ denote the input with length $T$ sampled from the video. It is fed into the SMS and FRP modules in parallel to capture the hierarchical spatial features and generate visual guidance maps. 

\begin{figure}[!htb]
  \centering
   \includegraphics[width=1.0\linewidth]{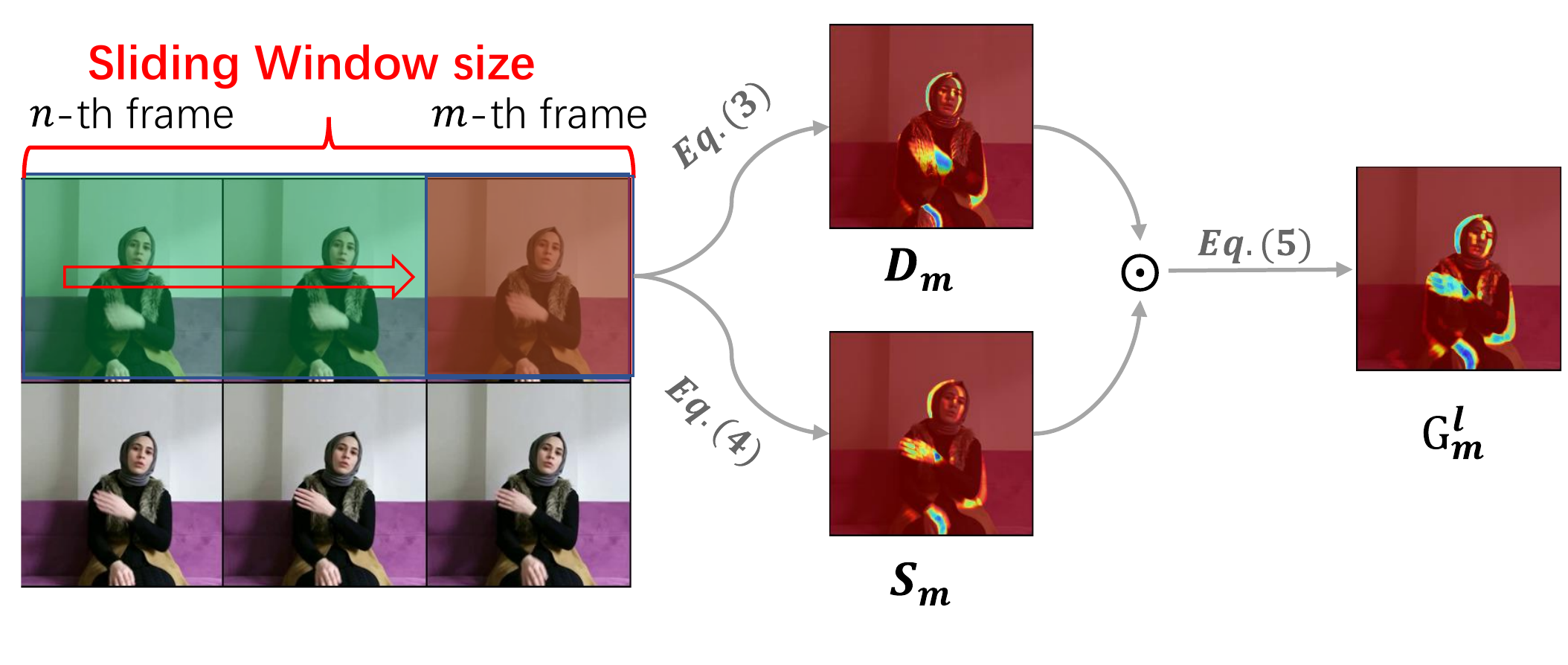}
   \caption{Overview of the proposed fast regional positioning module FRP. In each sliding window, the FRP module locates important areas in the last frame according to the successive video frames before it.
   }
   \label{fig:frp}
\end{figure}
\noindent\textbf{SMS Module.}\quad The SMS module is composed of a space-centric 3D Inception Module\footnote{The size of the convolution kernel in the spatial dimension is $1 \times 1$.} \cite{carreira2017quo} and a Max Pooling layer. It extracts the multi-scale features of $m$-th frame at the first layer by:
\begin{equation}\label{Eq:odt}
f_m^l = \textrm{Maxpool}(\mathcal{C}_{S-Inc}(I_m, W)), \quad s.t. \quad l=1
\end{equation}
where $l=1$ means it is the first layer of the DSN network; $\mathcal{C}_{S-Inc}(\cdot, W)$ indicates the Inception Module with learnable parameter matrix $W$; $\textrm{Maxpool}$ is the Max Pooling layer. Meanwhile, to guide each SMS module to focus on local important areas in the image, a visual guidance map is embedded in parallel to it to further enhance its visual perception, which is the proposed FRP module.

\noindent\textbf{FRP Module.}\quad  The FRP module is designed to generate the visual guidance map to rapidly locate important areas in an image for motion recognition. The visual guidance map is obtained directly by combining static and dynamic guidance maps. 
Specifically, as shown in Figure~\ref{fig:frp}, within a sliding window from the $n$-th to the $m$-th frame, we first calculate the dynamic image $DI(n,m)$ \cite{Zhou_Li_Wan_2021}:
\begin{equation}\label{Eq:dyimage}
\begin{split}
DI(n,m)=&DI(n-1,m-1)+(m-n)\times \\ 
&(I_{n-1}+I_{m})-2\sum\limits_{l=n}^{m-1} I_l, \quad s.t. \quad m>n
\end{split}
\end{equation}
where $DI(n-1,m-1)$ represents the dynamic image from the previous sliding window. Then the dynamic guidance map on $m$-th frame can be obtained by:
\begin{equation}\label{Eq:DIn}
D_m = \delta(DI(n,m))\times \lambda
\end{equation}
where $\delta$ and $\lambda$ indicate activation function and signal amplification factor respectively, herein we use GELU \cite{hendrycks2016gaussian} and $\lambda=2$ in all of experiments. 
However, we find that $D_m$ is sensitive to lighting as it only considers the motion information between multiple frames. To mitigate this problem, we introduce the static guidance map $S_m$, which can be defined as:
\begin{equation} \label{Eq:SIn}
\begin{split}
&S_m = \textrm{dilate}(\textrm{erode}(\hat{D}_{\rm{m}}), (k\times k))) \\
&\hat{D}_{\rm{m}}=
\begin{cases}
D_{\rm{m,(i,j)}} & D_{\rm{m,(i,j)}} >= D_{\rm{m,mean}}\\
0 & otherwise
\end{cases}
\end{split}
\end{equation}
where $\textrm{dilate}(\textrm{erode}(\cdot, (k\times k)))$ indicates the dilation and erosion operations with a kernel size of $k\times k$ in binary mathematical morphology.
$\hat{D}_{\rm{m}}$ is an attention matrix, in which all elements except for these higher than the mean value are set to zero, as we empirically observe that the response value at the region affected more by lighting is generally below average.
After that, combining the Eq.\ref{Eq:DIn} and \ref{Eq:SIn}, the visual guidance map of $m$-th frame at the $l$-th layer of the network can be derived by:
\begin{equation}\label{Eq:RPmap}
G^l_m = \textrm{Maxpool}((D_m + S_m)\times S_m)
\end{equation}
where the Max Pooling operation is used to scale the size of the current guidance map $G^{l}_m$ to match the feature map $f^l_m$. 
In addition, we also perform the normalization and alignment operations for the guidance map $G^{l}_m$, which have been discussed in detail in the supplementary material.
Finally, the spatial features $O \in \mathbb{R}^{T \times d}$ with length $T$ and dimension $d$ output from the DSN can be simply formulated as:
\begin{equation}\label{Eq:odt}
\begin{split}
& O = [O^l_1, O^l_2, \dots, O^l_T], \quad \forall l=1,2,\dots, M\\
& s.t. \quad O^l_m = (f^l_m \odot G_{m}^l)\otimes f^l_m + f^l_m
\end{split}
\end{equation}
where $M$ represents the number of total network layers used in DSN; $\odot$ and $\otimes$ represent element-wise product and 1D convolution operations respectively. 
It is then enhanced via RCM module for decoupled temporal representation learning.

\begin{figure}[!htb]
  \centering
   \includegraphics[width=1.0\linewidth]{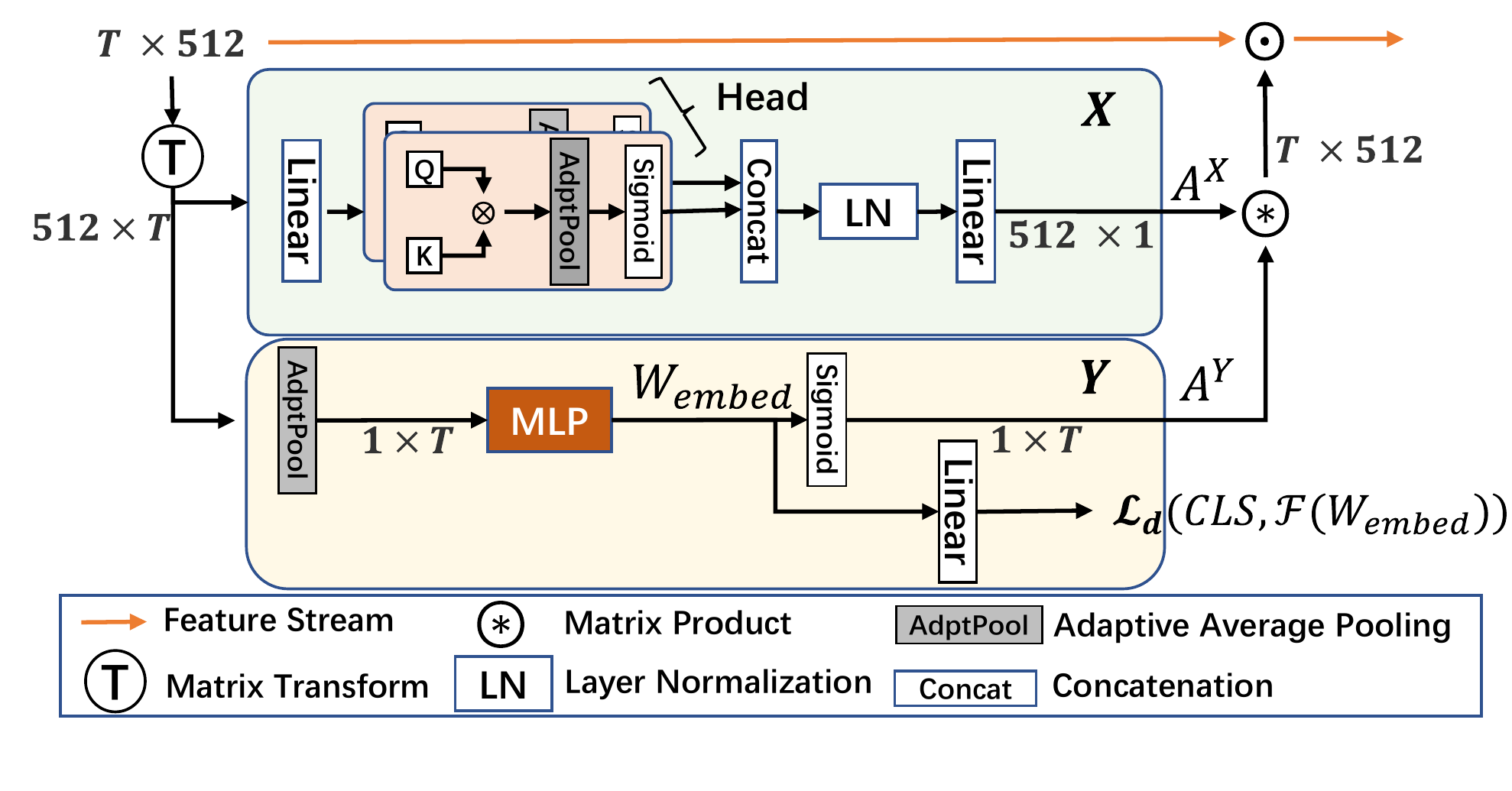}
   \caption{Overview of the proposed recoupling module RCM.}
   \label{fig:recoupling}
\end{figure}

\subsubsection{Decoupled Temporal Representation Learning (DTN)}
The temporal representation learning network DTN takes the \textbf{enhanced spatial features} $\hat O  \in \mathbb{R}^{T \times d/2}$ (formal definition in Sec.\ref{sec:recoupling}) as the input.
As shown in Figure~\ref{fig:pipline}, the DTN is configured as a multi-branch and two-stage structure to progressively learn the hierarchical temporal representation at local fine-grained level and global coarse-grained level. Specifically, for a single sub-branch $k$, we first sample a sub-sequence of features $\hat{O}_k \in \mathbb{R}^{T_n \times d/2}$ of length $T_n$ from $\hat{O}$ by a discrete sampling strategy:
\begin{equation}
\label{Eq:sample}
\begin{small}
\hat{O}_k = \{\hat{O}_\tau|\tau = \mathcal{R}[\lceil \frac{T}{T_n}\rceil \times t - 1, \lceil \frac{T}{T_n}\rceil \times t], t=1, 2, \dots, T_n \}
\end{small}
\end{equation}
where $\mathcal{R}$ is a random number generator. $\hat{O}_k$ serves as the input of temporal multi-scale features learning module (TMS).

\noindent\textbf{TMS Module.}\quad The TMS is composed of a time-centric 3D Inception Module\footnote{The size of the convolution kernel in the time dimension is 1.} $\mathcal{C}_{T-Inc}(\cdot, W_k)$ and Max Pooling layer.
We only use one layer TMS to capture the local fine-grained  spatial features $\hat{O}^L_k$:
\begin{equation}\label{Eq:local-fine}
\hat{O}^L_k = \textrm{Maxpool}(\mathcal{C}_{T-Inc}(\hat{O}_k, W_k))
\end{equation}
After that, $\hat{O}^L_k$ is fed into stack of Transformer blocks to learn the coarse-grained temporal representation.

\noindent\textbf{Transformer block.}\quad To reduce the redundant marginal information in captured temporal features, we utilize a Transformer structure based on $k$-NN multi-head self-attention layer~\cite{wang2021kvt}. Thus the modeling process of each Transformer block can be formulated as:
\begin{equation}\label{Eq:global-coarse}
\hat{O}^{G}_k = \textrm{MLP}(\textrm{LN}(\textrm{MSA}_{kNN}(\hat{O}^L_k))) + \hat{O}^{G-1}_k
\end{equation}
where $ \hat{O}^{G-1}_k$ indicates the output feature of the previous layer; $\textrm{MSA}_{kNN}(\cdot)$ indicates the $k$-NN multi-head self-attention layer and $\textrm{LN}$ represents layer normalization. 
Furthermore, to avoid overfitting to one of the sub-branches, we introduce a temperature parameter $\tau$ to control the sharpness of the output distribution of each sub-branch and impose a constraint loss on it. Therefore, the output of the DTN network can be formulated as:
\begin{equation}\label{Eq:sample}
\begin{small}
O^{CLS} = \sum_{k=1}^{K}{\textrm{MLP}(O^{CLS}_k) / \tau}, \forall k=1,2,3,\dots, K \\
\end{small}
\end{equation}
where $O^{CLS}$ is the class token vector embedded in the Transformer block and $K$ indicates the number of sub-branches. $\tau$ follows a cosine schedule from 0.04 to 0.07 during the training. 
We analytically demonstrate in Sec.\ref{sec:components} that the tactics of $k$-NN Attention, sharpness and multi-loss can bring performance gains for the proposed network. 

Based on spatial features $O$ from DSN and temporal features $O^{CLS}$ from DTN, we propose a recoupling strategy to strengthen the space-time connection during training, because the spatiotemporal decoupled learning method weakens the original coupled structure.

\subsection{Recoupling Spatiotemporal Representation}
\label{sec:recoupling}
The recoupling strategy is developed to rebuild the space-time interdependence, which reversely applies the distilled inter-frame correlations from time domain into the space domain through a self-distillation-based inner loop optimization mechanism during the training. 
Specifically, the spatial feature stream $O \in \mathbb{R}^{T\times d}$ derived from the spatial features learning network DSN is first linearly mapped into a low-dimensional space. Then the mapped features $\bar O \in \mathbb{R}^{T\times d/2}$ is transposed and fed into the recoupling module RCM.

\noindent\textbf{RCM module.}\quad The RCM module is configured as dual pathway to enhance the spatial features from the X (feature dimension) and Y (sequence dimension) directions, as shown in Figure~\ref{fig:recoupling}. For X direction (intra-frame), inspired by \textit{self-attention} \cite{dosovitskiy2020image}, we utilize a set of learnable matrices: $W_Q\in \mathbb{R}^{T\times T}$ and $W_K\in \mathbb{R}^{T\times T}$ to calculate the attention map $A_X$.
\begin{equation}\label{Eq:X}
\begin{split}
& Q=\bar{O}^TW_Q,\quad K= \bar{O}^TW_K, \\
& A_X = \delta(\textrm{GAP}(\frac{QK^T}{\sqrt{d}})), \quad A_X \in \mathbb{R}^{1\times d/2}
\end{split}
\end{equation}
where $Q$ and $K$ denote the queries and keys, $\delta$ is the Sigmoid activation function, and $\textrm{GAP}$ indicates the adaptive global average pooling operation. The attention map $A_X$ describes the correlation of the intra-frame in the sequence. 
For Y direction (inter-frame), the $\bar O$ is first compressed along the channel dimension through $\textrm{GAP}$ operation to obtain a feature vector with a dimension of $1 \times T$. Then it passes through an MLP block with two hidden layers to obtain a weight embedding $W_{embed}$.
\begin{equation}\label{Eq:Wembed}
W_{embed} = \textrm{MLP}(\textrm{GAP}(\bar{O}^T)), \quad W_{embed} \in \mathbb{R}^{1\times T}
\end{equation}
Finally, it is combined with a Sigmoid activation function to describe the correlation of the inter-frame in the sequence. 
\begin{equation}\label{Eq:Y}
A_Y = \delta(W_{embed}), \quad A_Y \in \mathbb{R}^{1\times T}
\end{equation}

However, the MLP layer cannot effectively learn the inter-frame correlation from the captured spatial features $\bar O$. Therefore, we employ a self-distillation loss function $\mathcal{C}_d$ to introduce additional supervision for the MLP block to distill inter-sequence correlation knowledge from temporal domain into $W_{embed}$. The distillation process can be formulated as:
\begin{equation}\label{Eq:distill}
\begin{split}
&\mathcal{C}_d = \mathcal{L}_d(CLS, \mathcal{F}(W_{embed})), CLS =  \sum_{k=1}^{K}{O^{CLS}_k}\\
& s.t. \quad \mathcal{L}_d(x, y) =\frac{1}{N_B} \sum_{i=1}^{N_B}{KL(x_i/\mathcal{T}-y_i/\mathcal{T})}
\end{split}
\end{equation}
where $\mathcal{T}$ is the distillation temperature parameter, $N_B$ is the batch size, $KL$ indicates Kullback-Leibler divergence \cite{hinton2015distilling} and $\mathcal{F}$ indicates linear mapping function. 
Combining Eq.\ref{Eq:X} and \ref{Eq:Y}, the attention map for spatial feature enhancement can be derived as:
\begin{equation}\label{Eq:w}
\begin{split}
A_{XY} = \sum_{i=1}^{I}{\sum_{j=1}^{J}{A^T_{X, i}\times A_{Y, j}}}, A_{XY}\in \mathbb{R}^{T\times d/2}
\end{split}
\end{equation}
where $I$ and $J$ indicate the element index in $A_X$ and $A_Y$ respectively. It is then applied to the raw spatial feature stream $O$ by the element-wise product operation:
\begin{equation}\label{Eq:enhance}
\hat{O} = O \odot A_{XY}
\end{equation}
where $\hat{O}$ represents the enhanced spatial feature, which is used for temporal representation learning.

\begin{figure}[t]
  \centering
   \includegraphics[width=1\linewidth]{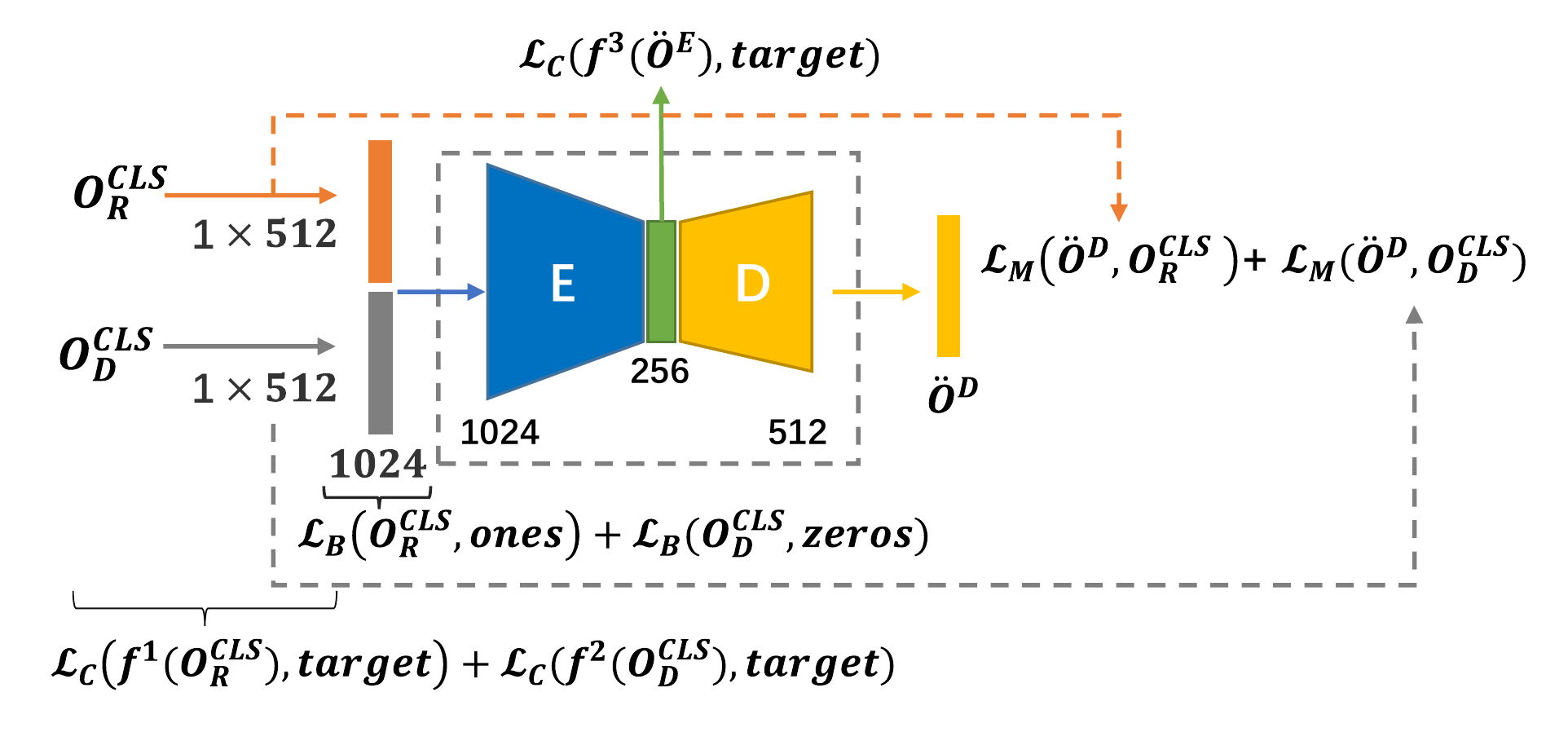}
   \caption{Adaptive fusion module CAPF.  $\mathcal{L}_C$, $\mathcal{L}_B$, $\mathcal{L}_M$ indicate cross entropy, binary cross entropy and mean square error loss functions, respectively;  $traget$ represents ground-truth; E and D indicate the Encoder and Decoder respectively.}
   \label{fig:adptfusion}
\end{figure}

\subsection{Cross-modal Interactive Learning}
\label{sec:fusion}
For the RGB-D multi-modal spatiotemporal features separately extracted from the two network branches, 
we propose to interact them at the spatial and temporal dimensions respectively at first, as shown in Figure~\ref{fig:CAPFM}.
Take the RGB modality as an example, the two spatial features from the RGB-D modalities are first interacted through the MLP layer to generate a joint spatial representation. It is then integrated with raw spatial features of the RGB modality by a residual structure:
\begin{equation}
    \ddot{O}^S_R = \textrm{LN}(\textrm{MLP}([\hat{O}_R^S||\hat{O}_D^S])) + O_R^S
\end{equation}
where $\hat{O}^S_R$ and $\hat{O}^S_D$ denote the enhanced spatial features of RGB-D modalities respectively; And $||$ represents concatenation. 
Similar to the interactive way of spatial features, after obtaining a joint temporal representation $\ddot{O}^T \in \mathbb{R}^{1\times d/2}$, it is fed into the cross-modal adaptive posterior fusion (CAPF) module for deep multi-modal features fusion. 

\noindent\textbf{CAPF module.}\quad
As shown in Figure~\ref{fig:adptfusion}, the CAPF module contains an Encoder and Decoder, which are composed of MLP blocks with multiple hidden layers. 
The $\ddot{O}^T$ is first fed into the Encoder to generate the encoded embedding $\ddot{O}^E \in \mathbb{R}^{1\times d/4}$ which can be used for classification. Then $\ddot{O}^E$ passes through the Decoder to obtain a decoded embedding $\ddot{O}^D \in \mathbb{R}^{1\times d/2}$ which can be used to supervise the Encoder. Therefore, the final classification score of the entire network can be obtained by:
\begin{equation}
\begin{small}
C = \textrm{Argmax}(f^1(O^{CLS}_R) + f^2(O^{CLS}_D) + f^3(\ddot{O}^E))
\end{small}
\end{equation}
where $f^*$ denotes the linear classifier and $\textrm{Argmax}$ is used to take the index of the maximum value in score vector.
Meanwhile, considering the challenge of optimizing the MLP layer, the entire network is trained by a multi-loss collaborative optimization strategy as shown in Figure~\ref{fig:adptfusion}.

\section{Experiments}
\label{sec:experim}

\subsection{Implementation Details}
The proposed method is implemented with Pytorch. The input sequences are spatially resized to $256\times256$, and then randomly cropped into $224\times224$ during training. The inputs are cropped into $224\times224$ around the center during inference.  We employ SGD as the optimizer with the weight decay of 0.0003 and momentum of 0.9. The learning rate is linearly ramped up to 0.01 during the first 3 epochs, and then decayed with a cosine schedule~\cite{loshchilov2016sgdr}. The training lasts for  100 epochs. The data augmentation only includes random clipping and rotation. Similar to \cite{yu2021searching}, all of our experiments except NTU-RGBD are pre-trained on 20BN Jester V1 dataset\footnote{https://www.kaggle.com/toxicmender/20bn-jester}. 
Moreover, three sub-branches are configured in DTN. The number of spatial and temporal feature learning blocks in DSN and DTN are $M=6$ and $N=6$, respectively.
We refer to this setting as the basic configuration of our network, unless otherwise specified. 

\subsection{Comparison with State-of-the-art Methods}
The proposed method achieves state-of-the-art performance on four gesture and action datasets. 
It is noted that we only list some of the state-of-the-art methods for comparison, and more comparison results can be found in the supplementary material.

\begin{table}[h]
  \centering
  \begin{tabular}{@{}lcc@{}}
    \toprule
    Method & Modality & Accuracy(\%) \\
    \midrule
    Transformer~\cite{d2020transformer} & RGB & 76.50\\
    MTUT~\cite{Abavisani_2019_CVPR} & RGB & 81.33\\
    NAS~\cite{yu2021searching} & RGB & 83.61\\
    Ours & RGB & \textbf{89.58}\\
    \bottomrule
    Transformer~\cite{d2020transformer} & Depth & 83.00\\
    MTUT~\cite{Abavisani_2019_CVPR} & Depth & 84.85\\
    NAS~\cite{yu2021searching} & Depth & 86.10\\
    Ours & Depth & \textbf{90.62}\\ 
    \bottomrule
    Transformer~\cite{d2020transformer} & RGB+Depth & 84.60\\
    MTUT~\cite{Abavisani_2019_CVPR} & RGB+Depth & 85.48\\
    MMTM ~\cite{Joze_2020_CVPR} & RGB+Depth & 86.31\\
    MMTM ~\cite{Joze_2020_CVPR} & RGB-D+Flow & 86.93\\
    PointLSTM~\cite{min2020efficient} & point clouds & 87.90\\
    NAS~\cite{yu2021searching} & RGB+Depth & 88.38\\
    \textbf{human} & RGB+Depth & 88.40\\
    \bottomrule
    Ours(Multiplication) & RGB+Depth & 90.89\\
    Ours(Addition) & RGB+Depth & 91.10\\
    Ours(CAPF) & RGB+Depth & \textbf{91.70}\\
    \bottomrule
  \end{tabular}
  \caption{Comparison of the state-of-the-art methods on the NvGesture.}
  \label{tab:sotanv}
\end{table}
\subsubsection{NvGesture Dataset}
The NvGesture \cite{molchanov2016online} dataset focuses on human-car interaction. It in total contains 1532 dynamic gesture videos (1050 for training and 482 for testing) in 25 classes. As can be seen in Table~\ref{tab:sotanv}, the proposed method significantly boosts performance (RGB: $\uparrow5.97\%$, depth: $\uparrow 4.52\%$ and RGB-D: $\uparrow3.3\%$) on this dataset for both single- and multi-modal configuration, which demonstrates its generalization ability in the field of human-computer interaction and small dataset. This might be because spatiotemporal decoupled modeling method prevents overfitting of the network to some extent.

\subsubsection{THU-READ Dataset}
The THU-READ \cite{tang2018multi} dataset consists 1920 videos with 40 different actions performed by 8 subjects. This dataset is challenging due to small-scale and background noise. For a fair comparison with other state-of-the-art methods, we follow the released leave-one-split-out cross validation protocol utilized in \cite{li2021trear}. As illustrated in Table \ref{tab:sotathu}, our method exceeds the state-of-the-art results and achieves the best average accuracy (87.40\%) in this protocol, which further demonstrates that our method is also robust to complicated background. We conjecture that this is mainly attributed to the proposed FRP module.

\begin{table}[h]
  \centering
  \begin{tabular}{@{}lcc@{}}
    \toprule
    Method & Modality & Accuracy(\%) \\
    \midrule
    SlowFast~\cite{feichtenhofer2019slowfast} & RGB & 69.58\\
    NAS~\cite{yu2021searching} & RGB & 71.25\\
    Trear~\cite{li2021trear} & RGB & 80.42\\
    Ours & RGB & \textbf{81.25}\\
    \bottomrule
    SlowFast~\cite{feichtenhofer2019slowfast} & Depth & 68.75\\
    NAS~\cite{yu2021searching} & Depth & 69.58\\
    Trear~\cite{li2021trear} & Depth & 76.04\\
    Ours & Depth & \textbf{77.92}\\
    \bottomrule
    SlowFast~\cite{feichtenhofer2019slowfast} & RGB+Depth & 76.25\\
    NAS~\cite{yu2021searching} & RGB+Depth & 78.38\\
    Trear~\cite{li2021trear} & RGB+Depth & 84.90\\
    \bottomrule
    Ours(Multiplication) & RGB+Depth & 86.10\\
    Ours(Addition) & RGB+Depth & 86.25\\
    Ours(CAPF) & RGB+Depth & \textbf{87.04}\\
    \bottomrule
  \end{tabular}
  \caption{Comparison of the state-of-the-art methods on the THU-READ.}
  \label{tab:sotathu}
\end{table}

\subsubsection{IsoGD Dataset}
The Chalearn IsoGD \cite{wan2016chalearn} dataset contains 47,933 RGB-D gesture videos divided into 249 kinds of gestures and is performed by 21 individuals. 
It is a much harder dataset because (1) it covers gestures in multiple fields and different motion scales from subtle fingertip movements to large arm swings, and (2) many gestures have a high similarity. However, as shown in Table \ref{tab:sotaIso}, our method also performs well on this dataset, possibly due to the hierarchical and compact features learned from the multi-scale network with the recoupling structure that can capture subtle differences from similar gestures.
\begin{table}[h]
  \centering
  \begin{tabular}{@{}lcc@{}}
    \toprule
    Method & Modality & Accuracy(\%) \\
    \midrule
    3DDSN~\cite{duan2018unified} & RGB &46.08 \\
    \makecell{AttentionLSTM}~\cite{zhu2019redundancy}  & RGB & 57.42 \\
    NAS~\cite{yu2021searching} & RGB & 58.88\\
    Ours & RGB & \textbf{60.87}\\
    \bottomrule
    AttentionLSTM~\cite{zhu2019redundancy}  & Depth & 54.18 \\
    3DDSN~\cite{duan2018unified} & Depth &54.95 \\
    NAS~\cite{yu2021searching} & Depth & 55.68\\
    Ours & Depth & \textbf{60.17}\\
    \bottomrule
    \makecell{AttentionLSTM}~\cite{zhu2019redundancy}  & RGB+Depth & 61.05 \\
    NAS~\cite{yu2021searching} & RGB+Depth & 65.54\\
    \bottomrule
    Ours(Multiplication) & RGB+Depth & 66.71\\
    Ours(Addition) & RGB+Depth & 66.68\\
    Ours(CAPF) & RGB+Depth & \textbf{66.79}\\
    \bottomrule
  \end{tabular}
  \caption{Comparison of the state-of-the-art methods on the Chalearn IsoGD.}
  \label{tab:sotaIso}
\end{table}

\subsubsection{NTU-RGBD Dataset}
The NTU RGB-D \cite{shahroudy2016ntu} is a large-scale human action dataset, which contains more than 56,000 multi-view videos of 60 actions performed by 40 subjects. This dataset is challenging due to large intra-class and viewpoint variations. As the skeleton information is available, many recent works tend to perform 2D/3D skeleton-based action recognition on this dataset since skeleton inherently highlights the key information of human body, whilst being robust to various illuminations and complex backgrounds. 
However, the generalization ability and robustness of skeleton-based methods are limited.
In this paper, we only use the modalities of color and depth for action recognition. As shown in Table~\ref{tab:ntu}, we achieve the state-of-the-art performance on both released protocols: Cross-view (CV) and Cross-subject (CS). 
\begin{table}[h]
  \centering
  \resizebox{1\linewidth}{!}{
  \begin{tabular}{@{}lccc@{}}
    \toprule
    Method & Modality & CS(\%) & CV(\%)\\
    \midrule
    Directed-GNN~\cite{shi2019skeleton} & Skeleton & 89.9 & 96.1\\
    Shift-GCN~\cite{2020Skeleton} & Skeleton & 90.7 & 96.5\\
    DC-GCN+ADG ~\cite{cheng2020decoupling} & Skeleton & 90.8 & 96.6\\
    CTR-GCN ~\cite{chen2021channel} & Skeleton & 92.4 & 96.8\\
    \bottomrule
    Chained Multi-stream ~\cite{2017Chained} & RGB & 80.8 & -\\
    SLTEP~\cite{2017The} & Depth & 58.2 & -\\
    DynamicMaps+CNN ~\cite{2018Depth} & Depth & 87.1 & 84.2\\
    DSSCA-SSLM~\cite{7892950} & RGB+Depth & 74.9 & -\\
    Cooperative CNN ~\cite{2017Cooperative} & RGB+Depth & 86.4 & 89.1\\
    Deep Bilinear ~\cite{Hu_2018_ECCV} & RGB-D+Skeleton & 85.4 & 90.7\\
    P4Transformer ~\cite{Fan_2021_CVPR} & point & 90.2 & 96.4\\
    \bottomrule
    Ours & RGB & 90.3 & 95.4\\
    Ours & Depth & 92.7 & 96.2\\
    Ours(Multiplication) & RGB+Depth & 93.6 & 96.6 \\
    Ours(Addition) & RGB+Depth & 93.9 & 96.7\\
    Ours(CAPF) & RGB+Depth & \textbf{94.2} & \textbf{97.3}\\
    \bottomrule
  \end{tabular}
  }
  \caption{Comparison of the state-of-the-art methods on the NTU-RGBD.}
  \label{tab:ntu}
\end{table}
Comparing with the skeleton-based state-of-the-art method CTR-GCN \cite{chen2021channel}, the proposed method achieves about 2\% improvement on CS protocol and 0.5\% on CV protocol. The performance of depth modality can be on par or even better, which further demonstrates the robustness of our method to noisy background and its strong motion perception abilities.

\section{Ablation Study}
NvGesture and THU-READ(CS2) are employed for the ablation study. All of experiments are conducted on RGB data modality except Sec.\ref{cmii}. We refer the reader to supplementary material for more ablation studies.

\subsection{Impacts of Embedded Components} 
\label{sec:components}
In Table~\ref{tab:ablacompo}, we show the impacts by introducing different components on the proposed network. First, we observe that in the absence of FRP module, the network does not work well either on NvGeture or THU-READ datasets, which shows that the early guidance of the network to focus on some local significant regions is beneficial to prevent the model from being trapped into the local optimum. We also demonstrate its robustness to lighting by visualization in supplementary material.
In addition, imposing a constraint to each branch of the network can prevent the model from overfitting to one of the branches. Sharpening  of the output distribution can encourage each sub-branch to learn more discriminative features. Finally, removing some of redundant information through $k$-NN Attention in the temporal representation can also bring certain performance gains.

\subsection{Effect of Recoupling Representation Learning}
\begin{figure}[t]
  \centering
  \includegraphics[width=1\linewidth]{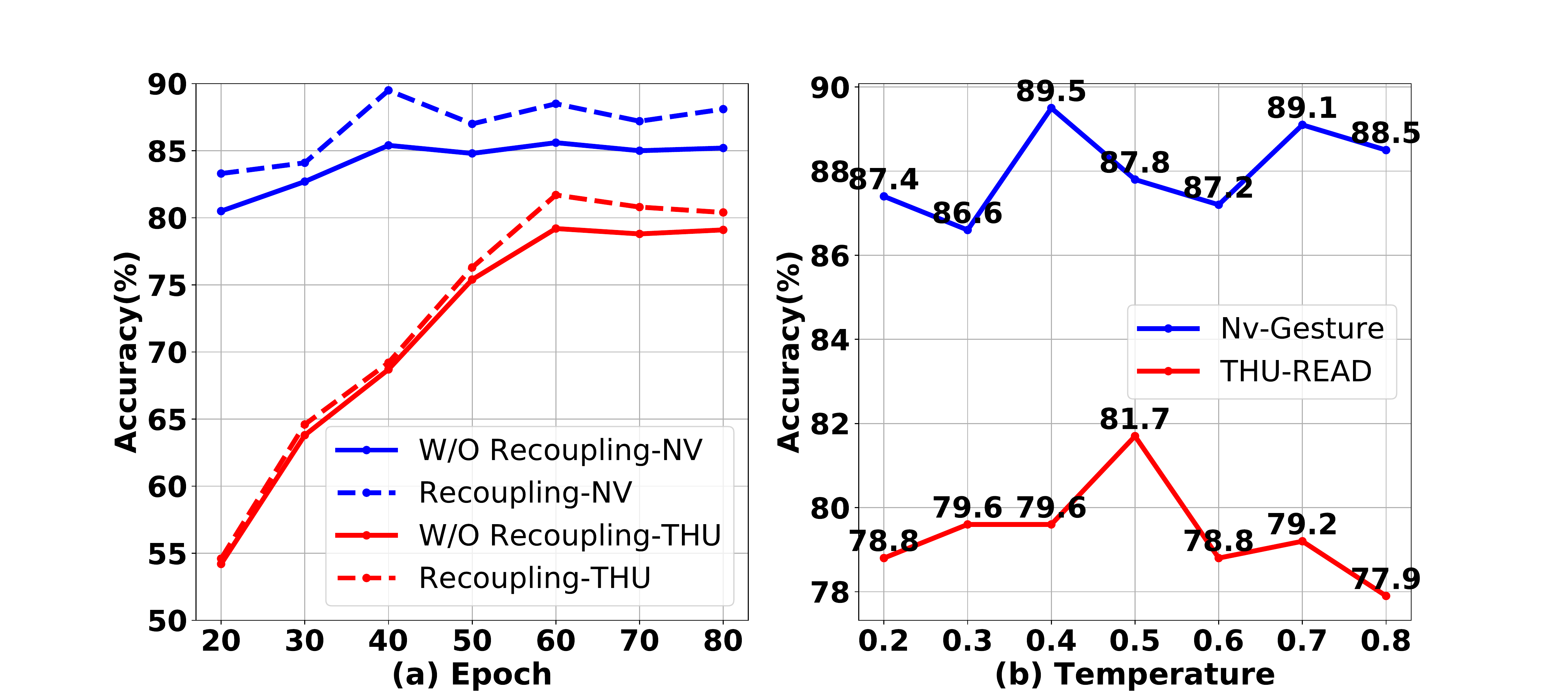}
  \caption{The ablation study of the recoupling strategy. (a) The effect of self-distillation-based recoupling learning on network performance. (b) The effect of distillation temperature on network performance.}
  \label{fig:recoup-temp}
\end{figure}
As shown in Figure~\ref{fig:recoup-temp} (a), recoupling learning can boost the performance (Nv: $\uparrow 3\%$, THU: $\uparrow 2\%$) and make the network converge rapidly. This is because distilling the knowledge from the temporal domain to enhance the spatial representation can help the network focus more on the informative features in the early stage of training. The idea behind it may be that the inner loop optimization mechanism based on self-distillation can help the network deviate from the local optima as soon as possible and move toward the global optimal solution. Figure~\ref{fig:recoup-temp} (b) shows the influence of distillation temperature $\mathcal{T}$ on network performance. We find that an appropriate temperature can boost the performance of the network. During the training stage, we set distillation temperature $\mathcal{T}$ as 0.4 and 0.5 for gesture and action datasets respectively.
\begin{table}[h]
  \centering
  \resizebox{1\linewidth}{!}{
  \begin{tabular}{@{}lccccccc@{}}
    \toprule
        \multirow{2}{*}{FRP} & \multirow{2}{*}{Multi-loss} & \multirow{2}{*}{Sharpness} & \multirow{2}{*}{\makecell{$k$-NN \\ Attention}} & \multicolumn{2}{c}{Accuracy(\%)} \\
        \cline{5-6} & & & & Nv & THU \\
    \midrule
    $\times$ & $\times$ & $\times$ & $\times$ &85.21 & 75.24\\ 
    \checkmark & $\times$ & $\times$ & $\times$ & 86.67 & 78.33\\
    \checkmark & \checkmark & $\times$ & $\times$ & 87.08 & 80.53\\
    \checkmark & \checkmark & \checkmark & $\times$ & 89.13 & 80.91\\
    \checkmark & \checkmark & \checkmark & \checkmark & \textbf{89.58} & \textbf{81.67}\\
    \bottomrule
  \end{tabular}
  }
  \caption{Impacts of some introduced components. The Multi-loss means that we impose constraint loss on each sub-branch in DTN.}
  \label{tab:ablacompo}
\end{table}

\subsection{Cross-modal Information Interaction}\label{cmii}
As shown in Table~\ref{tab:cross}, the communication of spatial or temporal information can boost the performance, wherein the information exchange at the spatial level alone can also bring performance gains (Nv: $\uparrow 0.2\%$, THU: $\uparrow 0.9\%$), which reflects that the interaction of cross-domain knowledge can help learn discriminative information. The last experiment shows that the joint effective cross-domain information interaction at the spatial and temporal levels can bring the largest performance gains (Nv: $\uparrow 0.6\%$, THU: $\uparrow 3.7\%$), verifying that multi-modal feature learning can benefit from spatiotemporal independent information transformation.
\begin{table}[h]
  \centering
  \begin{tabular}{@{}c|cccc@{}}
    \toprule
        Dataset & Baseline (Add)  & CmSI & CmTI & CmSTI \\
    \midrule
    NvGesture & 91.10 & 91.32 & 91.53 & 91.70\\
    THU-READ & 86.35 & 87.34 & 88.75 & 90.00 \\
    \bottomrule
  \end{tabular}
  \caption{Effects of cross-modal spatiotemporal information interaction. CmSI: Cross-modal Spatial information Interaction only. CmTI: Cross-modal Temporal information Interaction only.  CmSTI: Cross-modal spatiotemporal information Interaction.}
  \label{tab:cross}
\end{table}

\section{Conclusion}
We propose a method for unimodal decoupling and recoupling learning as well as a cross-modal interactive learning and fusion. Among them, we think that spatiotemporal recoupling learning is more important because it can lower the optimization difficulty,  especially under settings of  small-scale datasets.
In addition, we demonstrate that guiding the network to focus on the local important areas helps boost the performance. 
Finally, we find that interacting cross-modal spatiotemporal information at the spatial and temporal levels, respectively, can encourage the network to extract and fuses multi-modal spatiotemporal features. 

\section{Acknowledgment}
This work was supported by Alibaba Group through Alibaba Research Intern Program, the Science and Technology Development Fund of Macau (0008/2019/A1, 0010/2019/AFJ, 0025/2019/AKP, 0004/2020/A1, 0070/2020/AMJ), Key R\&D Programme of Guangdong Provincial (2019B010148001), the National Key Research and Development Plan under Grant (2021YFE0205700), the External cooperation key project of Chinese Academy Sciences (173211KYSB20200002), the Chinese National Natural Science Foundation Projects (61876179, 61961160704).

{\small
\bibliographystyle{ieee_fullname}
\bibliography{egbib}
}

\clearpage
\section{Supplementary Material}
\subsection{Supplementary for FRP Module}
\noindent\textbf{Normalization for Visual Guidance Map.}\quad 
To further improve the numerical stability, we do the normalization for generated visual guidance maps $G^{l}_m$ (Eq.5 in main manuscript) as:
\begin{equation}\label{Eq:norm}
G_{\rm{m, norm}}^l = \frac{G^{l}_m-G^{l}_{\rm{m,min}}}{G^{l}_{\rm{m,max}}-G^{l}_{\rm{m,min}}}
\end{equation}
where $G^{l}_{\rm{m,min}}$ and $G^{l}_{\rm{m,max}}$ represent the maximum and minimum values in $G^{l}_m$, respectively; And $G_{\rm{m, norm}}^l$ represents the normalized visual guidance map. 

\noindent\textbf{Visual Guidance Map Alignment.}\quad 
To align the generated visual guidance maps with the input sequence, we shift it backwards along the time dimension by $m-n$ units, and the guidance map of the previous $m-n$ frames is filled with the zeros matrix. Therefore, the final visual guidance map can be formulated as:
\begin{equation}
\begin{split}
&\hat G^l_{\rm{norm}} = [G^{l}_{1, \rm{norm}}, \dots, G^{l}_{T, \rm{norm}}], \forall l=1,2,\dots, M \\
&s.t. \quad G_{\rm{t,norm}}^l=
\begin{cases}
G_{\rm{t-(m-n),norm}}^l & t>m-n\\
0 & otherwise
\end{cases}
\end{split}
\end{equation}
where $\hat G^l_{\rm{norm}}$ represents the aligned visual guidance map with the input sequence. It then integrates with spatial feature stream captured by the spatial multi-scale features learning module (SMS) and serves as the input to next layer of the network.

\subsection{Structure of the SMS and TMS Modules} 
\begin{figure}[t]
\centering
\begin{subfigure}{0.2\textwidth}
\includegraphics[width=\linewidth]{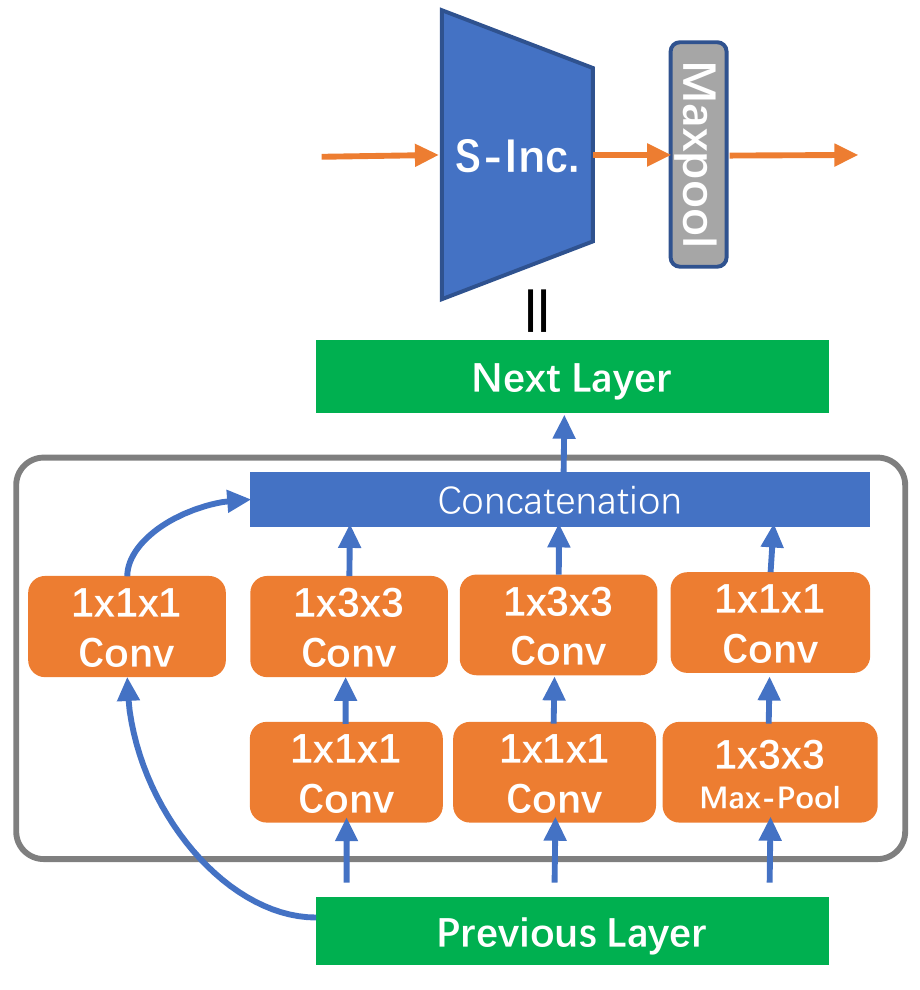}
\caption{SMS Module}
\label{fig:easy_aggregate}
\end{subfigure}
\begin{subfigure}{0.2\textwidth}
\includegraphics[width=\linewidth]{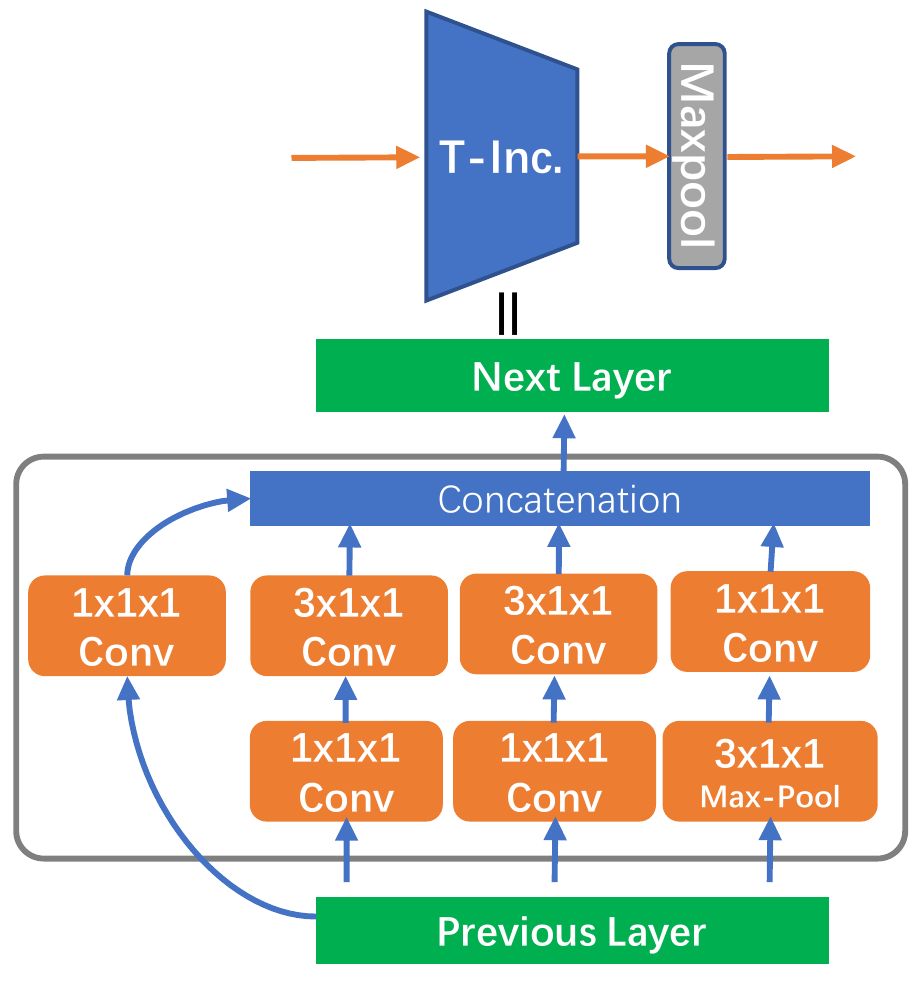}
\caption{TMS Module}
\label{fig:hard_aggregate}
\end{subfigure}
\caption{The structure of the spatial multi-scale features learning module SMS and the temporal multi-scale features learning module TMS.}
\label{fig:inception}
\end{figure}

As shown in Figure \ref{fig:inception}, the spatial and temporal multi-scale features learning module SMS and TMS are based on the inception structure. And a Max Pooling operation is embedded behind them to aggregate features with high correlation to reduce information redundancy.

\subsection{Loss Function}
For training the unimodal network, inspired by \cite{yu2021searching}, we configurate three sub-branches in the decoupled temporal representation learning network DTN, and each sub-branch imposes a constraint loss with weight coefficient of $\gamma$. 
In addition, we also introduce two additional constraint losses with weight coefficients of $1-\gamma$ and 1.0, to constrain the summation of three sub-branches and output of the RCM module.
So the overall loss for unimodal network training is the sum of all of those losses, and
can be denoted as: 
\begin{equation}
\begin{split}
    \mathcal{L}^{overall}_{\rm{uni}} = &\gamma \mathcal{L}^{\mathcal{S}_1}_C + \gamma \mathcal{L}^{\mathcal{S}_2}_C + \gamma  \mathcal{L}^{\mathcal{S}_3}_C + \\
    &(1-\gamma)  \mathcal{L}^{\mathcal{S}_{all}}_C + \mathcal{L}_D
\end{split}
\end{equation}
where $\mathcal{S}_1$, $\mathcal{S}_2$ and $\mathcal{S}_3$ represent the output of the three sub-branch respectively; $\mathcal{S}_{all} = \mathcal{S}_1 + \mathcal{S}_2 + \mathcal{S}_3$; and $\mathcal{L}_C$ and $\mathcal{L}_D$ represent classification loss and distillation loss, respectively.
For training the multi-modal network, we introduce a multi-loss collaborative optimization strategy, which can be denoted as:
\begin{equation}
\begin{split}
    \mathcal{L}^{overall}_{\rm{multi}} = &\mathcal{L}^{\mathcal{S}_R}_C + \mathcal{L}^{\mathcal{S}_D}_C + \mathcal{L}^{\mathcal{S}_R}_B + \mathcal{L}^{\mathcal{S}_D}_B + \\
    &\mathcal{L}^{\mathcal{S}_R}_M + \mathcal{L}^{\mathcal{S}_D}_M + \mathcal{L}^{\mathcal{S}_R}_D + \mathcal{L}^{\mathcal{S}_D}_D
\end{split}
\end{equation}
where $\mathcal{S}_R$ and $\mathcal{S}_D$ represent the output of the color and depth network branches, respectively;
and $\mathcal{L}_B$ and $\mathcal{L}_M$ represent binary cross entropy loss and mean square error loss.
It is note that we assign a weight coefficient of 1.0 to all losses. 

\subsection{Ablation Study}
\subsection{Impact of Sliding Window Size} 
\begin{table}[h]
    \centering
    \begin{tabular}{c|c|c|c|c|c|c}
    \toprule
          Size & 2 & 4 & 6 & 8 & 10 & 12 \\
         \midrule
          NvGesture & 87.5 & 87.7 & 88.1 & 88.8 & \textbf{89.6} & 88.2\\
          THU-READ & 80.4 & 80.8 & 80.8 & 81.2 & \textbf{81.7} & 80.6\\
         \bottomrule
    \end{tabular}
    \caption{The effect of the sliding window size.}
    \label{tab:winsize}
\end{table}

In Table \ref{tab:winsize}, we set different sliding window sizes in the FRP module to study how it affects network performance. We observe that the performance gradually improves as we increase the size of the window. However, when the size reaches 12, the performance of the network degrades instead. We conjecture that this may be because the response range in the dynamic guidance map has increased, and as a result, the value of some noise regions has also been amplified simultaneously.

\begin{figure*}[t]
\centering
\begin{subfigure}{0.3\textwidth}
\includegraphics[width=\linewidth]{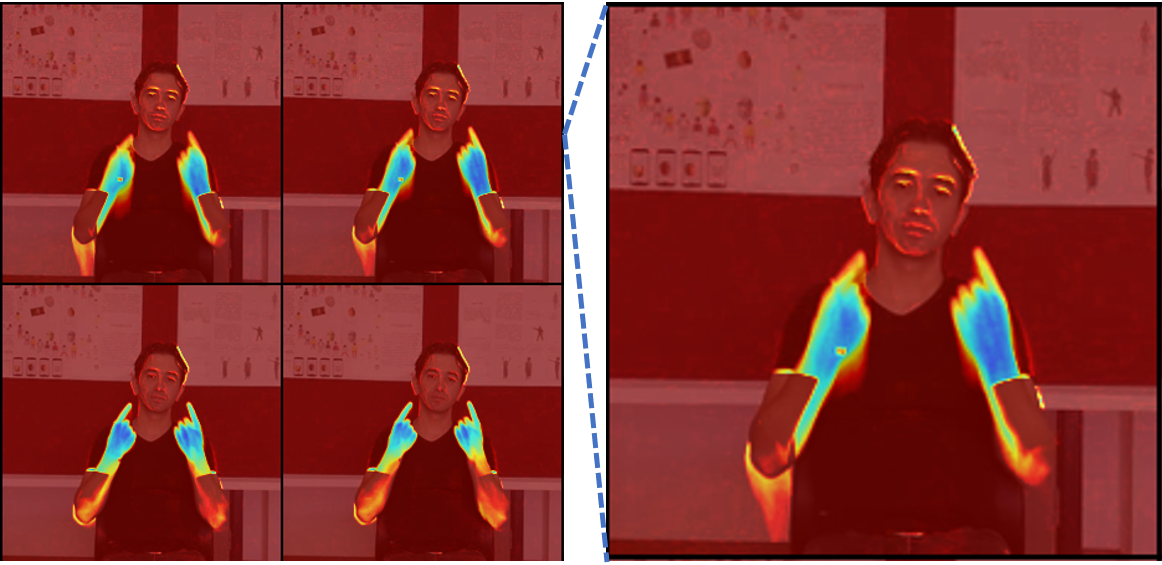}
\caption{Dynamic Guidance Map.}
\label{fig:easy_aggregate}
\end{subfigure}
\begin{subfigure}{0.3\textwidth}
\includegraphics[width=\linewidth]{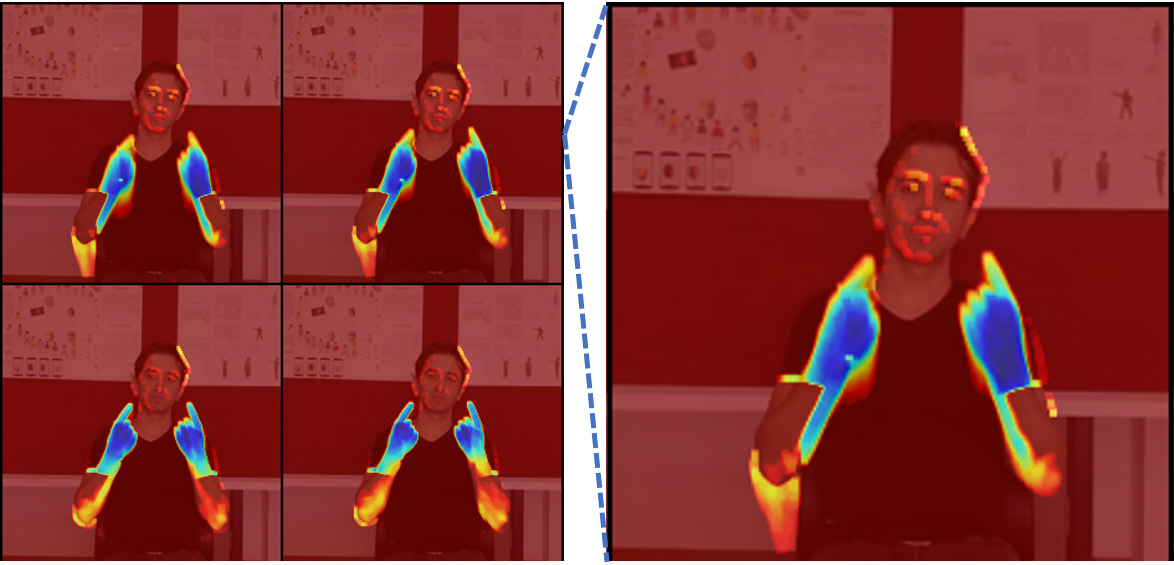}
\caption{Static Guidance Map.}
\end{subfigure}
\begin{subfigure}{0.3\textwidth}
\includegraphics[width=\linewidth]{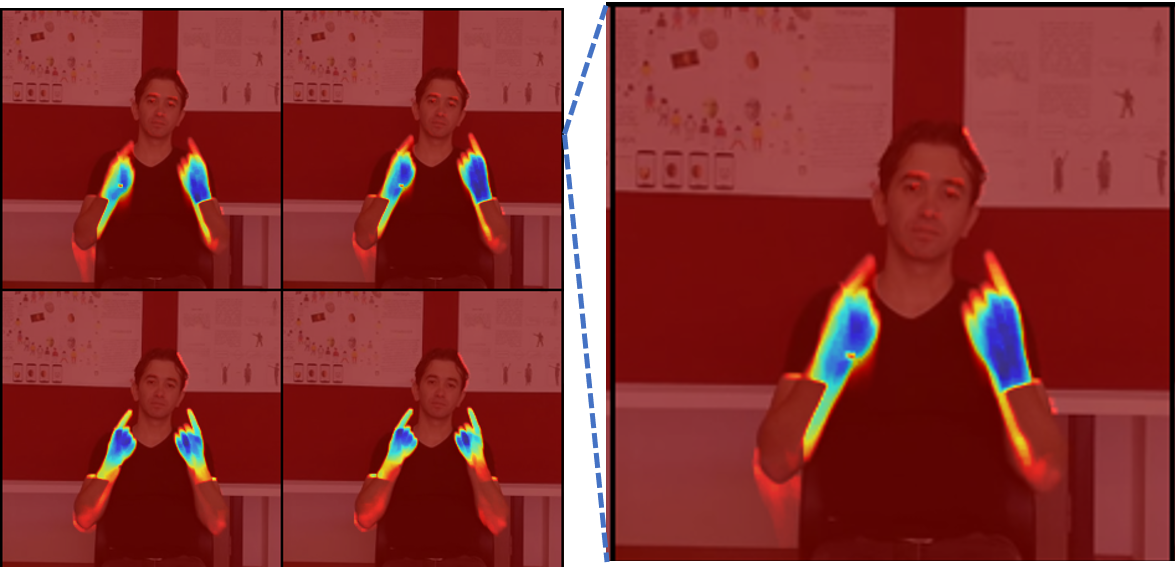}
\caption{Visual Guidance Map.}
\end{subfigure}
\caption{Visualization of the generated visual guidance map. (a) The dynamic guidance map defined with $D_m$ in the main manuscript. (b) The static guidance map defined with $S_m$ in the main manuscript. (c) The visual guidance map defined with $G_m$ in the main manuscript. Note that the deeper the color, the greater the weight.}
\label{fig:lighting}
\end{figure*}
\subsection{Study for the Robustness of Illumination} 
As shown in Figure \ref{fig:lighting} (a), the dynamic guidance map $D_m$ is inevitably influenced by illumination as it is driven by dynamic images. To address this issue, we introduce the static guidance map $S_m$, as shown in Figure \ref{fig:lighting} (b), it can not only enhance the response value of important areas in the image, but also significant alleviate the effects of lighting. After combining the dynamic guidance map and static guidance map, the final visual guidance map, as shown in Figure \ref{fig:lighting} (c), can effectively highlight the important areas in the image. 

\subsection{Impact of Local and Global Modeling in DTN}
\begin{figure}[t]
\centering
\begin{subfigure}{0.2\textwidth}
\includegraphics[width=\linewidth]{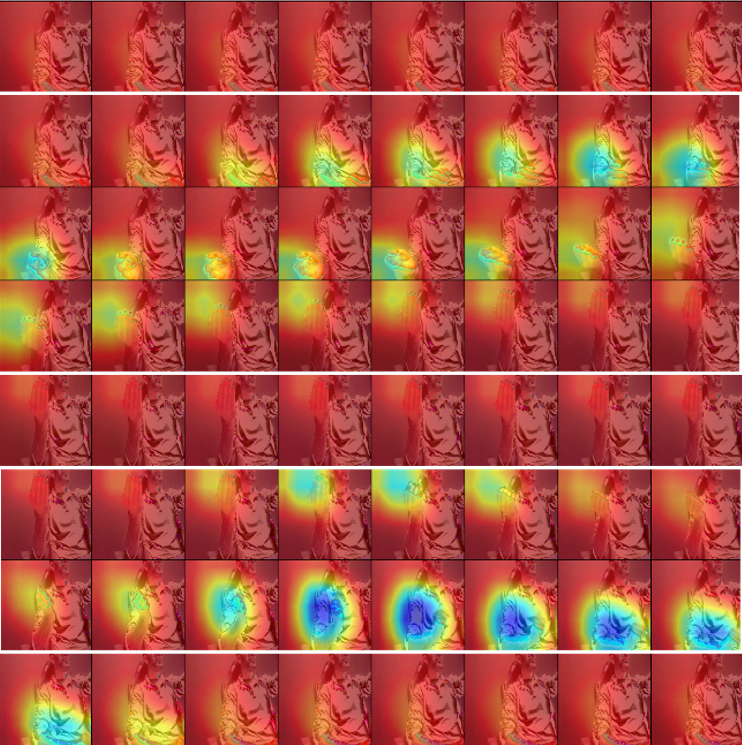}
\caption{Global Modeling.}
\label{fig:easy_aggregate}
\end{subfigure}
\begin{subfigure}{0.2\textwidth}
\includegraphics[width=\linewidth]{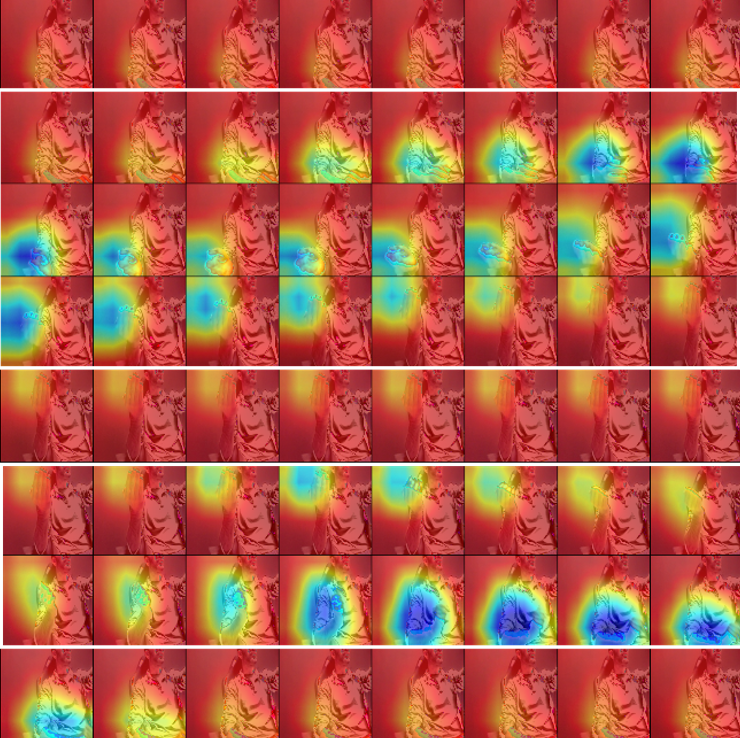}
\caption{Local and Global Modeling.}
\label{fig:hard_aggregate}
\end{subfigure}
\caption{Visualization of the Class Activation Map (CAM). (a) The activation response of global coarse-grained temporal information modeling. (b) The activation response of the joint modeling with local fine-grained as well as global coarse-grained temporal information.}
\label{fig:local-global}
\end{figure}
Temporal features learning based on global contextual information is vital for sequence. However, 
we find that solely utilizing the Transformer network for global contextual information modeling in the sequence is hard to generate effective motion descriptors, especially hard to capture the local subtle movement information as shown in Figure \ref{fig:local-global} (a). To alleviate this drawback, we introduce an inception-based temporal multi-scale features learning network (TMS) for local fine-grained temporal representation learning. It first captures local hierarchical temporal features, and then aggregates neighboring features with high correlation. After that, we feed them into stack of Transformer blocks to progressively learn the global temporal representation. As shown in Figure \ref{fig:local-global} (b), after modeling temporal information at a local fine-grained level and global coarse-grained level, the local and global motion perception abilities of the network have been significantly enhanced.

\subsection{Study for Feature Enhancement Attention}
\begin{figure}[t]
  \centering
  \includegraphics[width=0.9\linewidth]{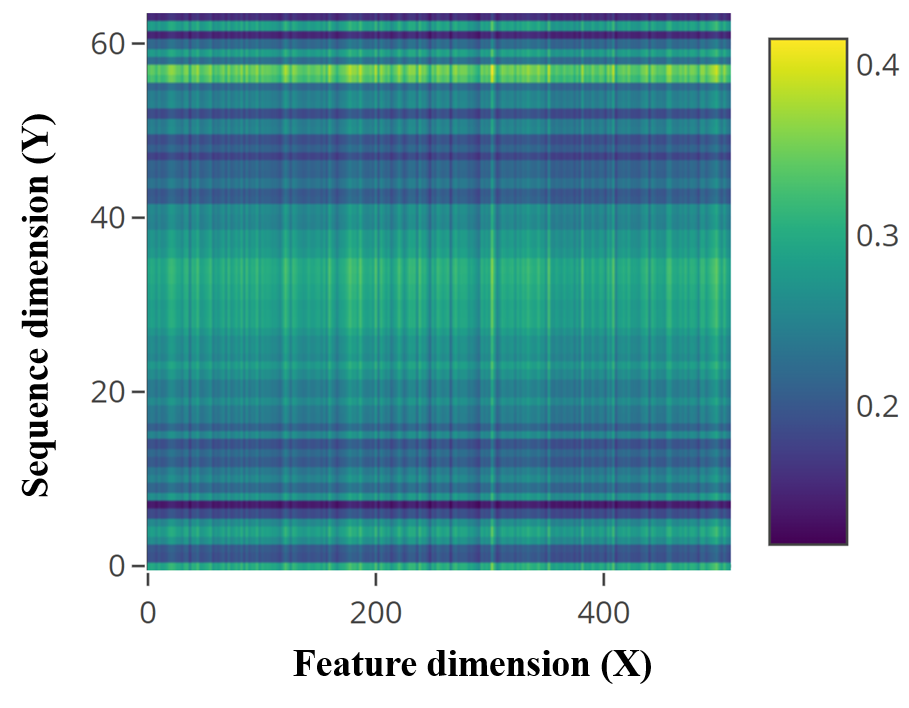}
  \caption{Visualization of the attention map for spatial feature enhancement generated by RCM module.}
  \label{fig:enhancement}
\end{figure}

Figure \ref{fig:enhancement} visualizes the attention map $A_{XY}$ (Eq.15 in main manuscript) generated by the spatiotemporal recoupling module (RCM), which shows that it can selectively activate several important neuron from X and Y directions in captured spatial features. 
In addition, we can obviously find that attention map $A_{XY}$ mainly guides the network to focus on the intermediate frame, which just shows that these intermediate frames contain most of the important information of a sequence. 

\subsection{Frame Rate Study for Sub-branch}
In this ablation, we configure different frame rates for each sub-branch to understand its impact on DTN. We only fine-tune the DSN sub-network and compare models trained for 100 epochs. As shown in Table \ref{tab:framerate}, the experiment result confirms that (1) configuring different frame rates for each sub-branch can boost the performance, which demonstrates that motion recognition benefits from multi-scale temporal features. And (2) setting a smaller or larger frame rate for DTN results in a decrease in performance, we conjecture that the former may be caused by the loss of important information, and the latter may be caused by temporal information redundancy. 
\begin{table}[h]
    \centering
    \begin{tabular}{ccc|cc}
    \toprule
        \makecell{Small \\ Transf.} & \makecell{Medium \\ Transf.} & \makecell{Large \\ Transf.} & Nv & THU \\
    \midrule
    16 & 16 & 16 & 79.88 & 77.08\\
    8 & 16 & 24 & 80.63 & 77.92\\
    16 & 32 & 48 & \textbf{81.46} & \textbf{78.75}\\
    16 & 48 & 80 & 81.30 & 77.92\\
    \bottomrule
    \end{tabular}
    \caption{The impact of different frame rates for each sub-branch in DTN. ``Transf." means Transformer network.}
    \label{tab:framerate}
\end{table}

\subsection{More Comparisons}
In this section, we compare with other methods not listed in the main manuscript.
Table \ref{tab:action} lists some other methods on the action datasets namely THU-READ and NTU-RGBD.
And Table \ref{tab:gesture} lists some other methods on the gesture datasets namely NvGesture and Chalearn IsoGD. 
\begin{table}[h]
  \centering
  \resizebox{1\linewidth}{!}{
  \begin{tabular}{@{}lccc@{}}
    \toprule
    \multicolumn{4}{c}{THU-READ Dataset} \\
    \midrule
    Method & Modality & \multicolumn{2}{c}{Accuracy(\%)} \\
    \midrule
    Appearance Stream~\cite{simonyan2014very} & RGB & \multicolumn{2}{c}{41.90}\\
    TSN~\cite{wang2016temporal} & RGB & \multicolumn{2}{c}{73.85}\\
    Ours & RGB & \multicolumn{2}{c}{\textbf{81.25}}\\
    \midrule
    Depth Stream~\cite{simonyan2014very} & Depth & \multicolumn{2}{c}{34.06}\\
    TSN~\cite{wang2016temporal} & Depth & \multicolumn{2}{c}{65.00}\\
    Ours & Depth & \multicolumn{2}{c}{\textbf{77.92}}\\
    \midrule
    MDNN~\cite{tang2018multi} & RGB+Flow+Depth & \multicolumn{2}{c}{62.92}\\
    TSN~\cite{wang2016temporal} & RGB+Flow & \multicolumn{2}{c}{78.23}\\
    TSN~\cite{wang2016temporal} & RGB+Flow+Depth & \multicolumn{2}{c}{81.67}\\
    Ours(CAPF) & RGB+Depth & \multicolumn{2}{c}{\textbf{87.04}}\\
    \midrule
    \multicolumn{4}{c}{NTU-RGBD Dataset} \\
    \midrule
    Method & Modality & CS(\%) & CV(\%)\\
    \midrule
    CNN+Motion+Trans~\cite{li2017skeleton} & Skeleton & 83.2 & 88.8\\
    ST-GCN~\cite{yan2018spatial} & Skeleton & 81.5 & 88.3\\
    Motif+VTDB~\cite{wen2019graph} & Skeleton & 84.2 & 90.2\\
    STGR-GCN~\cite{li2019spatio} & Skeleton & 86.9 & 92.3\\
    AS-GCN~\cite{li2019actional} & Skeleton & 86.8 & 94.2\\
    Adaptive GCN~\cite{shi2019two} & Skeleton & 88.5 & 95.1\\
    AGC-LSTM~\cite{si2019attention} & Skeleton & 89.2 & 95.0\\
    \midrule
    MMTM ~\cite{Joze_2020_CVPR} & RGB+Pose & 91.9 & -\\
    \midrule
    Ours & RGB & 90.3 & 95.4\\
    Ours & Depth & 92.7 & 96.2\\
    Ours(CAPF) & RGB+Depth & \textbf{94.2} & \textbf{97.3}\\
    \bottomrule
  \end{tabular}
  }
  \caption{Comparison with other methods on action datasets.}
  \label{tab:action}
\end{table}

\begin{table}[h]
  \centering
  \begin{tabular}{@{}lcc@{}}
    \toprule
    Method & Modality & Accuracy(\%) \\
    \midrule
     \multicolumn{3}{c}{NvGesture Dataset} \\
    \midrule
    GPM~\cite{GPM} & RGB & 75.90\\
    PreRNN~\cite{Yang_2018_CVPR} & RGB & 76.50\\
    ResNeXt-101~\cite{kopuklu2019real} & RGB & 78.63\\
    Ours & RGB & \textbf{89.58}\\
    \midrule
    ResNeXt-101~\cite{kopuklu2019real} & Depth & 83.82\\
    PreRNN~\cite{Yang_2018_CVPR} & Depth & 84.40\\
    GPM~\cite{GPM} & Depth & 85.50\\
    Ours & Depth & \textbf{90.62}\\ 
    \midrule
    PreRNN~\cite{Yang_2018_CVPR} & RGB+Depth & 85.00\\
    GPM~\cite{GPM} & RGB+Depth & 86.10\\
    Ours(CAPF) & RGB+Depth & \textbf{91.70}\\
    \midrule
    
    \multicolumn{3}{c}{Chalearn IsoGD Dataset} \\
    \midrule
    c-ConvNet~\cite{wang2018cooperative}   & RGB &36.60\\
    C3D-gesture~\cite{li2017large} & RGB & 37.28  \\
    AHL~\cite{hu2018learning}  &RGB & 44.88 \\
    ResC3D~\cite{miao2017multimodal} &RGB &45.07  \\
    3DCNN+LSTM~\cite{zhang2017learning}  & RGB &51.31 \\
    attention+LSTM~\cite{zhang2018attention}  & RGB & 55.98 \\
    Ours & RGB & \textbf{60.87}\\
    \midrule
    c-ConvNet~\cite{wang2018cooperative}    & Depth &40.08\\
    C3D-gesture~\cite{li2017large}  & Depth & 40.49  \\
    ResC3D~\cite{miao2017multimodal}  &Depth &48.44 \\
    AHL~\cite{hu2018learning} &Depth & 48.96  \\
    3DCNN+LSTM~\cite{zhang2017learning} & Depth &49.81 \\
    attention+LSTM~\cite{zhang2018attention}   & Depth & 53.28 \\
    Ours & Depth & \textbf{60.17}\\
    \midrule
    c-ConvNet~\cite{wang2018cooperative}    & RGB+Depth & 44.80\\
    AHL~\cite{hu2018learning}   &RGB+Depth & 54.14\\
    3DCNN+LSTM~\cite{zhang2017learning} &RGB+Depth & 55.29 \\
    Ours(CAPF) & RGB+Depth & \textbf{66.79}\\
    \bottomrule
    
  \end{tabular}
  \caption{Comparison with other methods on gesture datasets.}
  \label{tab:gesture}
\end{table}

\end{document}